\documentclass[journal]{IEEEtran}
\usepackage{cite}

\usepackage{amsmath,amssymb,amsfonts}
\usepackage{algorithmic}
\usepackage{gensymb}
\usepackage{amsthm}

\newtheorem{remark}{Remark}

\usepackage{graphicx,subfigure}
\usepackage{textcomp}
\usepackage{xcolor}
\usepackage{flushend}

\usepackage{subfigure}
\usepackage{pdfpages}
\usepackage{diagbox}
\usepackage{colortbl}
\definecolor{mygray}{gray}{0.85}
\usepackage{array}
\usepackage{caption}
\usepackage[linesnumbered,boxed,ruled,commentsnumbered]{algorithm2e}

\usepackage{eqparbox}

\usepackage{bbding}
\usepackage{tabularx}
\usepackage{tabularray}
\usepackage{booktabs}
\usepackage{amssymb}
\usepackage{xcolor}
\usepackage{array}
\ifCLASSINFOpdf
\else
\fi

\begin{document}
%
\title{Diffusion Models as Network Optimizers: Explorations and Analysis}
%
%
%

\author{Ruihuai~Liang,~\IEEEmembership{Student Member,~IEEE,} Bo~Yang,~\IEEEmembership{Member,~IEEE,} Pengyu~Chen, Xianjin Li, Yifan Xue, \\ Zhiwen Yu,~\IEEEmembership{Senior Member,~IEEE,} Xuelin Cao,~\IEEEmembership{Member,~IEEE,} Yan Zhang,~\IEEEmembership{Fellow,~IEEE},  M\'erouane Debbah,~\IEEEmembership{Fellow,~IEEE}, H. Vincent Poor,~\IEEEmembership{Life Fellow,~IEEE}, and Chau~Yuen,~\IEEEmembership{Fellow,~IEEE}  
\thanks{
This work was supported in part by the National Natural Science Fund
for Excellent Young Scientists Fund Program (Overseas) and in part by the Qin Chuang Yuan Fund Program under Grant QCYRCXM-2022-358. The work of Zhiwen Yu was supported in part by the National Natural Science Foundation of China under Grant 61960206008. The work of Xuelin Cao was supported in part by the Natural Science Basic Research Program of Shaanxi under Grant 2023-JC-QN-0650, in part by the Qin Chuang Yuan Fund Program under Grant QCYRCXM-2022-240, and in part by the Special Fund for Local Research and Development Program of Shaanxi under Grant 2023ZY1-CGZY-01. The work of H. V. Poor was supported in part by a grant from Princeton Language and Intelligence. The work of Chau Yuen was supported in part by the National Research Foundation Singapore under the AI Singapore Programme (AISG Award No: AISG2-TC-2023-008-SGKR).

R. Liang, B. Yang, P. Chen, X. Li, Y. Xue, and Z. Yu are with the School of Computer Science, Northwestern Polytechnical University, Xi'an, Shaanxi, 710129, China. X. Cao is with the School of Cyber Engineering, Xidian University, Xi'an, Shaanxi, 710071, China. Y. Zhang is with the Department of Informatics, University of Oslo, 0316 Oslo, Norway. 
M. Debbah is with  KU 6G Research Center, Department of Computer and Information Engineering, Khalifa University, Abu Dhabi 127788, UAE and also with CentraleSupelec, University Paris-Saclay, 91192 Gif-sur-Yvette, France.
H. V. Poor is with the Department of Electrical and Computer Engineering, Princeton University, Princeton, NJ 08544, USA.
C. Yuen is with the School of Electrical and Electronics Engineering, Nanyang Technological University, Singapore. 

Pengyu~Chen, Xianjin Li, and Yifan Xue contribute equally.

 \textit{Corresponding author:} Bo Yang (Email: yang$\_$bo@nwpu.edu.cn).}
}

%
%

\markboth{Journal of \LaTeX\ Class Files,~Vol.~14, No.~8, August~2015}%
{Shell \MakeLowercase{\textit{et al.}}: Bare Demo of IEEEtran.cls for IEEE Journals}
%


\maketitle

\begin{abstract}
Network optimization is a fundamental challenge in the Internet of Things (IoT) network, often characterized by complex features that make it difficult to solve these problems. Recently, generative diffusion models (GDMs) have emerged as a promising new approach to network optimization, with the potential to directly address these optimization problems. However, the application of GDMs in this field is still in its early stages, and there is a noticeable lack of theoretical research and empirical findings.
In this study, we first explore the intrinsic characteristics of generative models. Next, we provide a concise theoretical proof and intuitive demonstration of the advantages of generative models over discriminative models in network optimization. Based on this exploration, we implement GDMs as optimizers aimed at learning high-quality solution distributions for given inputs, sampling from these distributions during inference to approximate or achieve optimal solutions.
Specifically, we utilize denoising diffusion probabilistic models (DDPMs) and employ a classifier-free guidance mechanism to manage conditional guidance based on input parameters. We conduct extensive experiments across three challenging network optimization problems. By investigating various model configurations and the principles of GDMs as optimizers, we demonstrate the ability to overcome prediction errors and validate the convergence of generated solutions to optimal solutions. 
We provide code and data at https://github.com/qiyu3816/DiffSG.

\end{abstract}

\begin{IEEEkeywords}
Internet of things, network optimization, diffusion models, generative artificial intelligence. 
\end{IEEEkeywords}

%
\IEEEpeerreviewmaketitle

\section{Introduction}


\IEEEPARstart{N}{etwork} optimization problems arise from the widespread need to maximize or minimize specific objectives while utilizing limited network resources~\cite{shi2023survey}. In the Internet of Things (IoT) network, these problems vary in form depending on the scenario, aiming for optimal resource allocation with highly dense deployed IoT devices~\cite{iot-mac,iot1}. Common resources include communication capacity, storage, computational power, and energy. The optimization objectives typically involve maximizing transmission rates~\cite{noma2021uav}, minimizing total delay~\cite{iot-delay} and energy consumption~\cite{iot-energy,zeng2024online}, reducing execution costs~\cite{yang2021computation,feng2023resourceAll,liang2023globe}, or enhancing quality of service (QoS)~\cite{serviceCaching2020info,serviceCaching2021}. 
The complexity of real-world environments and application requirements introduces challenging mathematical properties to these problems, such as convexity and non-convexity, linearity and non-linearity, and solution spaces that may be discrete, continuous, or mixed-integer. Additionally, these problems can involve both differentiable and non-differentiable components. These properties impact not only the objective functions but also the constraints, resulting in a complex feasible solution space that makes solving such problems particularly challenging. 

Existing network optimization methods primarily include numerical approaches based on optimization theory~\cite{noma2021uav,zeng2024online,feng2023resourceAll,serviceCaching2020info,serviceCaching2021,sun2024multi,10382630} and fitting algorithms based on machine learning~\cite{yang2021computation,liang2023globe,jiang2020deeplearning,gnn2023zhang,gnn2022supervised,cao-jsac,async2021dai,teal2023xu,leo2024xie,multi2024zhao}, with some work exploring the use of deep learning to enhance numerical methods~\cite{learn2branch2020,learn2cut2024,AI-survey}. For problems without complex characteristics, it is often straightforward to apply classical algorithms. However, when dealing with more typical challenges—such as non-convex, mixed-integer, multi-objective, or Pareto optimization—custom algorithm design becomes necessary~\cite{noma2021uav,zeng2024online,feng2023resourceAll,serviceCaching2020info,serviceCaching2021,sun2024multi,10382630}. This process demands a strong understanding of optimization theory and can be labor-intensive in developing effective solutions, which makes IoT network designers burden. In the case of fitting algorithms, standard supervised learning \cite{yang2021computation,liang2023globe,jiang2020deeplearning,gnn2023zhang,gnn2022supervised} has limited capability to directly learn the mapping from input to optimal solutions, particularly when dealing with non-differentiable objective functions or constraints. Under such conditions, full convergence is nearly impossible, making it difficult to meet complex optimization demands. As a result, reinforcement learning (RL)~\cite{async2021dai,teal2023xu,leo2024xie,multi2024zhao}, which interacts with the solution space and environment to make decisions, has become more commonly used. However, RL performance is heavily dependent on extensive interaction and exploration, which makes it challenging to train in high-dimensional, complex solution spaces of real IoT networks. Additionally, RL often suffers from issues such as reliance on reward function design and a tendency to get trapped in local optima. In numerical methods enhanced by deep learning~\cite{learn2branch2020,learn2cut2024}, certain components of traditional algorithms that are difficult to construct manually are replaced by learning-based approaches. While this can improve performance on some problems, it also demands more specialized designs compared to conventional numerical methods and hinders the efficiency of IoT construction. In summary, \textit{while conventional network optimization methods have achieved notable success, they often come with the drawbacks of heavy exclusive designs and inherent limitations in learning capacity. } 
In this paper, we propose a novel approach focused on learning high-quality solution distributions for network optimization, which has the potential to overcome the above shortcomings. Accordingly, we introduce generative models for network optimization, especially diffusion models~\cite{diffusion2023survey,ddpm2020ho}, which have demonstrated strong performance in recent years. \vspace{-2mm}

\begin{figure}[t]
\centerline{\includegraphics[width=3.3in]{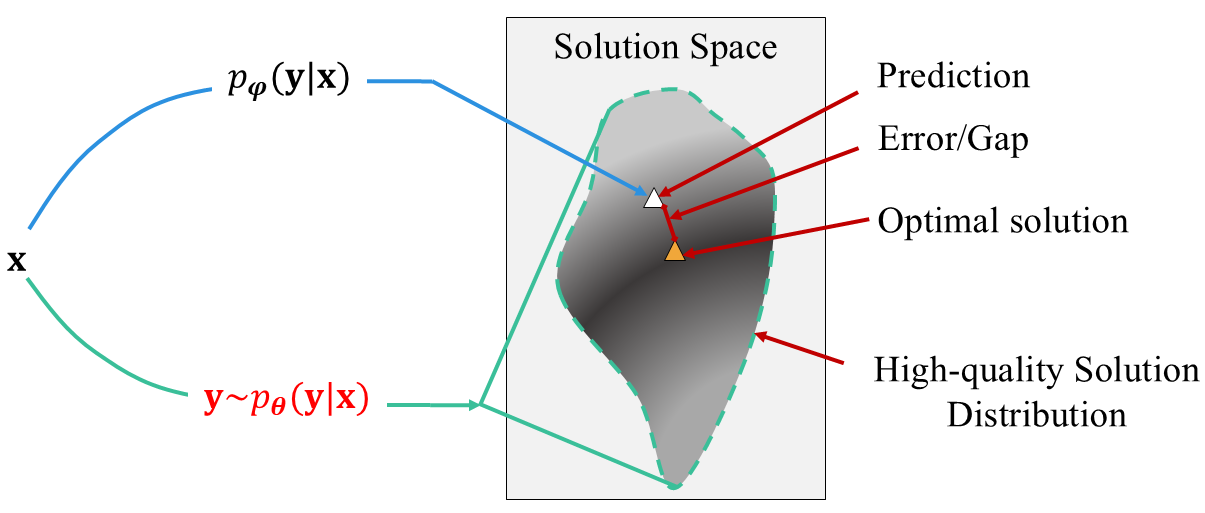}}
\caption{\small The learning results of generative models differ from those of discriminative models. Discriminative models focus on mapping $\mathbf{x}$ directly to the optimal solution, but they often incur learning errors that are difficult to eliminate. In contrast, generative models learn the distribution of high-quality solutions for a given input $\mathbf{x}$, with the optimal solution typically lying within this distribution.}
\label{fig_gen_advan}
\vspace{-4mm}
\end{figure}

\subsection{Our Proposal}
We first answer the intuitive question: ``\textit{Why use generative models for network optimization?}'' Recall that the key difference between generative and discriminative models is that generative models learn the joint probability distribution $P(\mathbf{x},\mathbf{y})$, while discriminative models learn the conditional probability distribution $P(\mathbf{y}|\mathbf{x})$. The core purpose of generative models is to capture a probability distribution and sample from it, whereas discriminative models focus on tasks like classification and regression. Additionally, generative diffusion models can leverage $P(\mathbf{y}|\mathbf{x})$ to guide the sampling of specific types within the learned distribution \cite{beat2021gans,ho2022classifier}. 

Current fitting algorithms in network optimization are predominantly discriminative models, which aim to map single-point or sequential inputs to optimal solutions. However, existing neural networks struggle to fit complex objective functions directly, and this inverse fitting process is even more constrained. As a result, there is often a discrepancy between the output solution and the true optimal value, as illustrated in Fig. \ref{fig_gen_advan}. Due to the complexity of the target high-quality solution distribution in most cases, this error is difficult to mitigate using random strategies in RL, and nearly impossible to eliminate in standard supervised learning. 

However, we find that generative models possess a global awareness of the entire solution space for a given input, along with the ability to repeatedly sample solutions by learning the joint probability distribution from input to output, as illustrated in Fig. \ref{fig_gen_advan}. Specifically, inspired by \cite{dimes2022qiu}, we define the conditional generation target of the generative diffusion model as a probability distribution over solutions, $p_{\theta}(\mathbf{y}|\mathbf{x})$, where $\mathbf{x}$ is the given input and better solutions have higher probabilities. In this way, the model captures the feasible solution space corresponding to the input $\mathbf{x}$ and probabilistically describes the quality of different solutions within it. Through repeated sampling, the model can approximate or even reach the global optimal solution, effectively closing the gap between traditional discriminative methods and the true optimal value. Notably, this approach to learning the solution distribution is insensitive to the complex characteristics of the original objective and constraints. 

\subsection{Contributions}
This work focuses on applying generative diffusion models (GDMs) to network optimization problems, providing both theoretical and experimental analysis to demonstrate the effectiveness of this emerging approach. Our key contributions are as follows: 
\begin{enumerate}
    \item We analyze the current state of network optimization methods in IoT and GDMs for optimization problems, answering the key question: ``\textit{why use generative models for network optimization?}'' We show that GDMs can understand the global solution space by learning the distribution of high-quality solutions to generate optimal results and provide theoretical discussions. 
    \item We demonstrate the effectiveness of GDMs as network optimizers across several typical IoT problems, visualizing the generation process to enhance understanding of how GDMs converge. 
    \item Given the limited application of GDMs as network optimizers directly, we conduct extensive exploratory experiments on the configuration of diffusion models. Our results explore applying GDMs to a broader range of optimization problems. 
\end{enumerate}
\vspace{-2mm}

\section{Literature Review and Analysis}

\renewcommand{\arraystretch}{1.5}
\begin{table*}[]
\caption{\small Comparison of various approaches utilizing GDMs for optimization and networking tasks.}
\label{tab_taxonomy}
\begin{tabularx}{\textwidth}{>{\centering\arraybackslash}m{0.14\textwidth}|>{\centering\arraybackslash}m{0.16\textwidth}|>{\centering\arraybackslash}m{0.18\textwidth}|>{\centering\arraybackslash}m{0.2\textwidth}|>{\centering\arraybackslash}m{0.22\textwidth}}
\hline
\textbf{Works} & \textbf{Is Optimizer} & \textbf{For Networking} & \textbf{With Mechanism Analysis} & \textbf{With Implementation} \\ \hline
\cite{difusco2023sun,t2t_2023,diff2024gasoline,li2024diffusionmodeldatadrivenblackbox,Krishnamoorthy2023BBO} & \textcolor{black}{\LARGE \textbf{\checkmark}} & \XSolidBrush & \XSolidBrush & \textcolor{black}{\LARGE \textbf{\checkmark}} \\
\hline
\cite{sun2024generativeaideepreinforcement,du2024D2SAC,liu2024deep,huang2023ai,wu2023cddm,rf2024diffusion,10759093} & \XSolidBrush & \textcolor{black}{\LARGE \textbf{\checkmark}} & \XSolidBrush & \textcolor{black}{\LARGE \textbf{\checkmark}} \\
\hline
\cite{du2023beyond,liang2024diffsggenerativesolvernetwork,wang2024generativeaibasedsecure} & \textcolor{black}{\LARGE \textbf{\checkmark}} & \textcolor{black}{\LARGE \textbf{\checkmark}} & \XSolidBrush & \textcolor{black}{\LARGE \textbf{\checkmark}} \\
\hline
\cite{haoran2024diff,yuan2024diff_drl,guo2024gradientguidancediffusionmodels} & \XSolidBrush & \XSolidBrush & \XSolidBrush & \textcolor{black}{\LARGE \textbf{\checkmark}} \\
\hline
\cite{letafati2023diffusion} & \XSolidBrush & \textcolor{black}{\LARGE \textbf{\checkmark}} & \XSolidBrush & \XSolidBrush \\
\hline
\textbf{Ours} & \textcolor{black}{\LARGE \textbf{\checkmark}} & \textcolor{black}{\LARGE \textbf{\checkmark}} & \textcolor{black}{\LARGE \textbf{\checkmark}} & \textcolor{black}{\LARGE \textbf{\checkmark}} \\
\hline
\end{tabularx}
\vspace{-3mm}
\end{table*}

\subsection{Diffusion Models}
Diffusion models are a class of generative models that operate by gradually adding noise to data in a forward process, and learn to reverse this process to remove the noise, ultimately recovering the original data. In the forward process, fresh data is progressively transformed into noise, while in the reverse process, the model learns to denoise the data corrupted on different levels, reconstructing the original data. This forward-reverse mechanism is closely tied to the Fokker-Planck and Langevin dynamics, which describe how probability distributions evolve under stochastic forces. The concept was first formalized by \cite{sohl2015deep}, who demonstrated that a carefully designed reverse diffusion process could recover high-quality data from noise. This concept was further implemented in the idea of score matching~\cite{before2019ddpm}. 

Diffusion models began to flourish after \cite{ddpm2020ho} proposed the breakthrough Denoising Diffusion Probabilistic Models (DDPMs), 
which have risen to prominence in the Generative Artificial Intelligence (GAI) community. DDPMs demonstrated the ability of these models to generate high-resolution, detailed samples across various domains, such as image synthesis, natural language processing, and audio generation~\cite{diffusion2023survey,tkde2024survey}. DDPMs introduce a two-step process: gradually adding Gaussian noise to the data as the forward process, and denoising the data step by step as the learnable reverse process. This enables GDMs to excel at reconstructing complex target data distributions from a known noise distribution. 

To generate target data distributions that align with a given prompt, rather than sampling randomly across a broad data distribution, \cite{beat2021gans} proposed classifier-guided diffusion generation. This approach uses a pre-trained image classifier to guide the generation toward images of the specified class. However, since this method relies on an explicit image classifier, \cite{ho2022classifier} introduced classifier-free guidance for diffusion generation. This alternative enables the diffusion model itself to incorporate an implicit classifier, achieving the same conditional generation without needing an external classifier. 

Furthermore, to address the concern of overly long denoising processes in GDMs, numerous accelerated sampling methods, such as Denoising Diffusion Implicit Models (DDIMs) \cite{song2020ddim} and Direct Optimization of Diffusion Latents (DOODL) \cite{Wallace_2023_end_end}, have been developed to reduce the total sampling steps without compromising generation quality. For non-image data, the theory and practice of diffusion generation in continuous, discrete, and structured data spaces have also become relatively mature \cite{austin2021structured,difusco2023sun}.

Finally, the applications of diffusion models have expanded well beyond their initial use in image generation, achieving widespread success in interdisciplinary research areas such as biology, chemistry, and materials science \cite{graphDiff2023survey}. This further demonstrates the GDM's exceptional generalization ability for complex distributions \cite{li2023generalization} and it is potential for applications across various structured data types \cite{graphDiff2023survey,diffusion2023survey}.

\subsection{Diffusion in Networking and Optimization}

Research on applying GDMs to network optimization is still in its early stages. We classify existing works according to the usage of GDMs and review their development history. 

\subsubsection{GDMs as Optimizers} To begin, we can consider two works \cite{difusco2023sun,t2t_2023} from the AI community that use graph diffusion models to address classical combinatorial optimization problems as the origins of applying GDMs as optimizers. Among these, DIFUSCO \cite{difusco2023sun} focuses on generating solutions for large-scale Traveling Salesman Problems (TSP) and Maximum Independent Set (MIS) problems, achieving stable state-of-the-art results in offline optimization. Built on DIFUSCO, T2T \cite{t2t_2023} addresses the lack of explicit integration of the optimization objective function in DIFUSCO’s conditional guidance by embedding the gradient of the objective function within the input conditioning, leading to performance improvements. While DIFUSCO and T2T have made remarkable theoretical and practical contributions, there remains room for improvement in terms of objective and constraint complexity. The TSP and MIS objectives are inherently differentiable and have relatively simple constraints, which do not fully match the complexity of typical network optimization problems. 

In the networking domain, only a limited number of studies have directly applied GDMs as optimizers. \cite{du2023beyond} provides a case study using GDMs for a convex optimization problem, although it focuses more on using GDMs to enhance RL. \cite{liang2024diffsggenerativesolvernetwork} provides an early explanation of the mechanisms and advantages of using GDM as a network optimization solver, along with exploratory validation on several common network optimization problems. \cite{wang2024generativeaibasedsecure} employs discrete diffusion to generate graphs that minimize operational costs and utilizes continuous diffusion to create signals that evade monitoring. In \cite{diff2024gasoline}, DIFUSCO’s approach is adapted to model Pareto optimization within Gantt chart representations, utilizing DDPM to generate Gantt charts for solution generation. However, as DIFUSCO argues, modeling problems as image representations may not enable explicit learning of the optimization problem itself. \cite{liang2024gdsggraphdiffusionbasedsolution} employs graph diffusion to solve the optimization problems in Multi-access Edge Computing (MEC) networks and explores the convergence of the denoising solution generation process, but the method itself still has limited generality problems. 

In addition, GDM has shown significant potential in a class of problems known as closed-box optimization \cite{li2024diffusionmodeldatadrivenblackbox,Krishnamoorthy2023BBO}. Closed-box optimization primarily addresses problems with unknown objective functions, often arising from AI for Science (AI4S). These problems tend to be highly complex, meaning a model is typically tailored for a specific optimization setup. This one-shot nature of the GDM optimizer contrasts with the versatility and efficiency required in network optimization scenarios. Furthermore, recent work has provided an intriguing reverse perspective, applying optimization theory to the implementation of GDM conditional generation \cite{guo2024gradientguidancediffusionmodels}. 


\subsubsection{GDMs for RL Enhancement} In fact, the use of GDMs to enhance RL also originates from the AI field \cite{haoran2024diff,yuan2024diff_drl}, primarily in offline RL. These efforts primarily leverage GDMs’ strong ability to learn multi-modal data distributions—where “multi-modal” here refers to the probabilistic definition, describing complex data distributions for which a probability density function is nearly impossible to express analytically. This highlights the capacity of diffusion models to capture intricate distributions, underscoring the necessity and rationale for using generative models to learn distributions rather than traditional fitting algorithms. 

In the network community, \cite{sun2024generativeaideepreinforcement} discusses RL enhanced by GAI. \cite{du2024D2SAC} proposes D2SAC, which applies diffusion models as agents in RL frameworks for network optimization. Similarly, \cite{liu2024deep} discusses how diffusion models can streamline wireless network management. Their framework focuses on generating optimal network configurations, but many theoretical and experimental questions remain open. \cite{huang2023ai} introduces a diffusion-based approach for AI-generated network design (AIGN), automating intricate design tasks and optimizing network performance without requiring human expertise. \cite{10759093} studies a multi-objective optimization problem of air communication security and energy efficiency using diffusion-enabled RL. 

\subsubsection{GDMs for Synthesis} Meanwhile, using GDMs purely as tools for feature extraction, data reconstruction, and synthesis remains the mainstream approach in the networking domain and continues to grow \cite{wu2023cddm,letafati2023diffusion,rf2024diffusion,wang2024generativeaibasedsecure}, which also relies on the multi-modal distribution learning ability of the GDMs.

\subsubsection{Other Generative Models as Optimizers}
Notably, solving optimization problems with generative models suggests that a wide range of generative methods have the potential to function as optimizers, an area currently under active research. This includes approaches that leverage prompt engineering or fine-tuning of Large Language Models (LLMs) for solution optimization \cite{yang2024largelanguagemodelsoptimizers,llm2024evol_opt,li2024large,mongaillard2024large,zou2024telecomgpt,lee2024largelanguagemodelsknowledgefree}. Specifically, \cite{yang2024largelanguagemodelsoptimizers} first uses the trajectory of previous iterations combined with external prompts to guide an LLM toward the final solution of a target optimization problem. And \cite{li2024large} introduces LLMs into multi-objective optimization, proposing an LLM-Enabled Decomposition-based Multi-objective evolutionary Algorithm (LEDMA) to optimize Uncrewed Aerial Vehicle (UAV) deployment and power control in integrated sensing and communication systems. Through careful prompt engineering to balance communication and sensing performance in non-convex problems, the model was guided to generate superior solutions iteratively. Similarly, \cite{llm2024evol_opt} explores using LLM-driven evolutionary algorithms to solve the classic TSP problem. \cite{mongaillard2024large} applies prompt engineering to transform user voice requests into optimization problems, developing an LLM-based multi-agent framework for power scheduling in electric vehicle charging networks. They demonstrate how LLMs, with the help of prompts, can adjust scheduling strategies in real-time based on user preferences, enhancing energy efficiency while maintaining quality of service. \cite{lee2024largelanguagemodelsknowledgefree} studies that LLMs can eliminate the dependency on prior knowledge and be seamlessly applied for various network management tasks. Additionally, \cite{zou2024telecomgpt} develops TelecomGPT, a specialized LLM framework for the telecommunications industry. By leveraging domain-specific datasets and fine-tuning prompts, TelecomGPT enhances its ability to tackle resource allocation and network management tasks, especially providing feasibility validation for translating natural language prompts into mathematical models.

\subsection{Summary}

As summarized in Table. \ref{tab_taxonomy}, existing works have achieved valuable results in the performance of GDM optimizers and GDM-assisted RL. However, there has been little comprehensive explanation for why GDM or GAI should be directly used as an optimizer. Although many studies can utilize conditional diffusion generation to achieve optimized solutions, the intrinsic randomness of GDMs raises questions among those accustomed to deterministic optimization methods. Our work addresses this gap by offering a preliminary, reasoned perspective on ``\textit{mechanism of GAI in optimization}". We aim to provide insights into the mechanisms, theory, and experimental guidance for using GDMs as network optimizers. \vspace{-2mm}

\section{Diffusion Model-empowered Network Optimizer}

\begin{figure*}[t]
\centering
\setlength{\abovecaptionskip}{-0.5mm}
\includegraphics[width=13 cm]{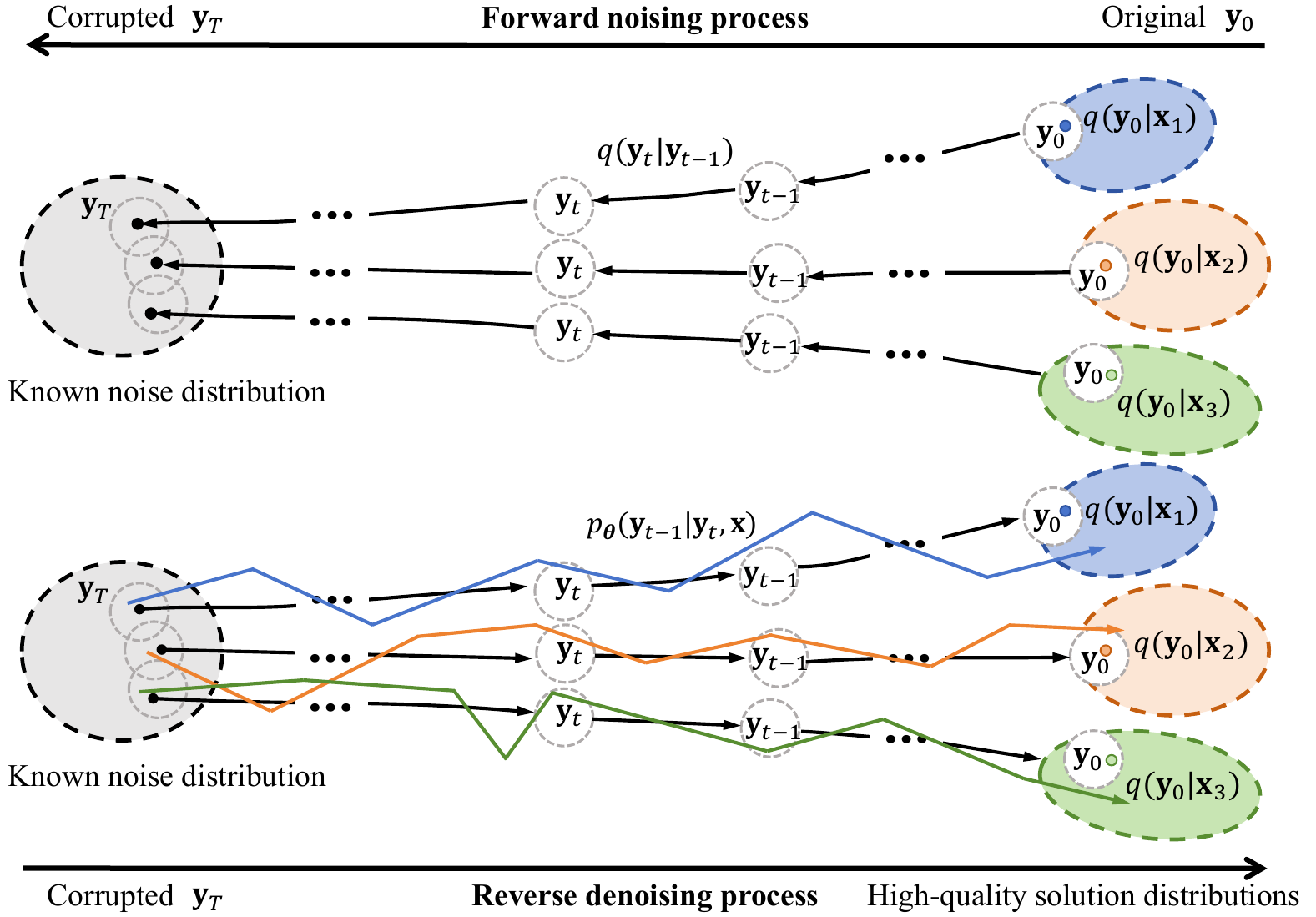}
\caption{\small  Forward noising process and reverse denoising process of GDM as an optimizer. The generation process involves randomness (represented by colored arrows) and ultimately converges to the target high-quality solution distribution.}
\label{fig_method_framework}
\end{figure*}

\subsection{Discussion of Generative Advantages}\label{sec_gen_advan}
A typical network optimization problem can be formalized as an optimization problem: $\min_{\mathbf{y} \in \mathcal{Y}} f(\mathbf{x}, \mathbf{y}), \mathbf{x} \in \mathcal{X}$, where $\mathbf{x}$ represents input parameters, $\mathbf{y}$ denotes optimization variables, $f\rightarrow\mathbb{R}$ is the objective function, and $\mathcal{X}$ and $\mathcal{Y}$ are the domain and feasible region, respectively. 

For this problem, we briefly demonstrate in this section the advantages of generative methods over discriminative methods, specifically that the expected objective function value of the output solutions has a lower bound. 

First, we define the discriminative model $\varphi$ and the generative model $\theta$. For a given $\mathbf{x}$, their outputs are $h_{\varphi}(\mathbf{x})$ and $\mathbf{y} \sim p_{\theta}(\mathbf{y}_0|\mathbf{x})$, respectively. Since model training cannot perfectly fit the target, we introduce a prediction error $\mathbf{e}$. For simplicity and generality, we make no assumptions about the data format, dimensions, or magnitude of this error. 

\subsubsection{Lower Bound of the Expected Objective Function Value of Discriminative Model} The above definition means that $h_{\varphi}(\mathbf{x})$ will deviate from the optimal solution $\mathbf{y}^*$ and the degree is determined by $\mathbf{e}$ 
\begin{equation}\label{eq_dis_pred_error}
    h_{\varphi}(\mathbf{x})\in\partial{\mathrm{U}}(\mathbf{y}^*,\mathbf{e}),
\end{equation}
where $\partial{\mathrm{U}}(\mathbf{y}^*,\mathbf{e})$ represents the boundary of the $\mathbf{e}$ neighborhood of $\mathbf{y}^*$. 

Assume there is a probability \( p \) that a solution outside \( \mathrm{U}(\mathbf{y}^*, \mathbf{e}) \) exceeds the objective function value of the optimal solution \( f(\mathbf{x}, \mathbf{y}^*) \) by a factor of $\sigma > 1$ 
\begin{equation}\label{eq_sigma_exceed_pr}
    \mathrm{Pr}\left(f(\mathbf{x},\mathbf{y}) > f(\mathbf{x},\mathbf{y}^*)\sigma\ \mathrm{and}\ \mathbf{y}\notin\mathrm{U}(\mathbf{y}^*,\mathbf{e})\right)=p,
\end{equation}
Now, we can derive the lower bound of the expected objective function value for the solution predicted by $\varphi$ 
\begin{equation}\label{eq_dis_obj_func_lb}
    \mathbb{E}\left[ f(\mathbf{x},h_{\varphi}(\mathbf{x})) \right]\geq pf(\mathbf{x},\mathbf{y}^*)\sigma+(1-p)f(\mathbf{x},\mathbf{y}^*),
\end{equation}
where the first term on the right side of the inequality represents cases where the predicted solution satisfies the probability \( p \) condition in Eq. \ref{eq_sigma_exceed_pr}, while the second term represents cases where it does not. In both terms, the objective function values reflect the best possible values under each probability scenario, making the right side of the inequality a valid lower bound. 

\subsubsection{Lower Bound of the Expected Objective Function Value of Generative Model} Intuitively, we consider that the prediction error implies that the predicted optimal solution with the highest probability in the high-quality solution distribution generated by the model will deviate from \( \mathbf{y}^* \) by an amount \( \mathbf{e} \). However, although the predicted solution distribution of the generative model is shifted, it generally still retains a certain sampling probability for the true optimal solution. Therefore, we denote \( p_i \) as the probability that a single sampled solution belongs to \( \mathrm{U}(\mathbf{y}^*, \mathbf{e}) \) 
\begin{equation}\label{eq_gen_pred_in_neighbor}
    \mathrm{Pr}\left( \mathbf{y}\sim p_{\theta}(\mathbf{y}_0|\mathbf{x})\ \mathrm{and}\ \mathbf{y}\in\mathrm{U}(\mathbf{y}^*, \mathbf{e}) \right)=p_i,
\end{equation}
Similarly, we denote \( p_o \) as the probability that a single sampled solution is out of \( \mathrm{U}(\mathbf{y}^*, \mathbf{e}) \) 
\begin{equation}\label{eq_gen_pred_out_neighbor}
    \mathrm{Pr}\left( \mathbf{y}\sim p_{\theta}(\mathbf{y}_0|\mathbf{x})\ \mathrm{and}\ \mathbf{y}\notin\mathrm{U}(\mathbf{y}^*, \mathbf{e}) \right)=p_o,
\end{equation}
where $p_i+p_o=1$. 

In line with Eq. \ref{eq_dis_obj_func_lb}, we can derive the lower bound of the expected objective function value for the solution sampled by the generative model 
\begin{equation}\label{eq_gen_obj_func_lb}
\begin{aligned}
\mathbb{E} & \left[ f(\mathbf{x},\mathbf{y}\sim p_{\theta}(\mathbf{y}_0|\mathbf{x})) \right] \geq p_i f(\mathbf{x},\mathbf{y}^*) \\
& \quad + p_o\left[pf(\mathbf{x},\mathbf{y}^*)\sigma+(1-p)f(\mathbf{x},\mathbf{y}^*)\right],
\end{aligned}
\end{equation}

\subsubsection{Conclusion} By subtracting the right side of Eq. \ref{eq_gen_obj_func_lb} from the right side of Eq. \ref{eq_dis_obj_func_lb} and substituting $p_i+p_o=1$ we obtain 
\begin{equation}\label{eq_gen_advan_conclusion}
    f(\mathbf{x},\mathbf{y}^*)(\sigma-1)p_i p>0. 
\end{equation}

\begin{remark}[better optimization bound of solution generation]\label{remark_advan}
Eq. \ref{eq_gen_advan_conclusion} implies that, under the same settings, the generative model can produce solutions that are superior to those of the discriminative model. This is primarily because, given the same prediction error, the discriminative model cannot correct or eliminate errors in its existing predictions to approach the optimal solution. In contrast, the generative model fundamentally learns the distribution of high-quality solutions, enabling it to sample different solutions and maintain the ability to approximate or reach the optimal solution, as illustrated in Fig. \ref{fig_gen_advan}. Thus, we highlight the advantages that the learning objectives of the generative model bring to the performance of optimization problems. 
\end{remark}

\subsection{Optimizer Formulation}
Based on the above definition, we explain the noising and denoising processes of the discrete-time DDPM adopted in our work to learn the distribution of high-quality solutions $p(\mathbf{y}_0|\mathbf{x})$, where the index 0 represents the original, uncorrupted data. Each pair $(\mathbf{x}, \mathbf{y})$, representing an input and its corresponding optimal solution, forms our training dataset $\mathcal{D}$. Since $\mathbf{x}$ has no influence during the noising process, we simplify this to $p(\mathbf{y}_0)$; it only serves as a fixed condition to guide the input during the denoising process. 

For the noising process, the original data is gradually corrupted by Gaussian noise with mean and variance controlled by the noise factor $(\alpha_t)^T_{t=1}$. Given an uncorrupted training sample $\mathbf{y}_0\sim p(\mathbf{y}_0)$, the noisy samples $\mathbf{y}_1,\mathbf{y}_2 ...,\mathbf{y}_T$ can be obtained through the following Markov process: 
\begin{equation}\label{eq_ddpm_noising}
    p(\mathbf{y}_t|\mathbf{y}_{t-1})=\mathcal{N}(\mathbf{y}_t;\sqrt{\alpha_t}\cdot \mathbf{y}_{t-1},(1-\alpha_t)\cdot \mathbf{I}),\ \forall t\in{1,...,T},
\end{equation}
where $T$ represents the number of diffusion steps. Additionally, $\mathbf{I}$ stands for the identity matrix with the same dimension as $\mathbf{y}_0$.

However, acquiring corrupted data $\mathbf{y}_t$ for any step by totally depending on Eq. \ref{eq_ddpm_noising} necessitates an excessive number of noising iterations. Consequently, leveraging the recursive nature of this formula allows for the direct sampling of $\mathbf{y}_t$ using the simplified approach outlined as 
\begin{equation}\label{eq_ddpm_onestep_noising}
p\left(\mathbf{y}_t \mid \mathbf{y}_0\right)=\mathcal{N}\left(\mathbf{y}_t ; \sqrt{\bar{\alpha}_t} \cdot \mathbf{y}_0,\left(1-\bar{\alpha}_t\right) \cdot \mathbf{I}\right),
\end{equation}
where $t$ is drawn from a uniform distribution $\mathcal{U}(\{1,...,T\})$, and $\bar{\alpha}_t$ is defined as the product of ${\alpha}_i$ for $i$ ranging from $1$ to $t$. Consequently, Eq. \ref{eq_ddpm_onestep_noising} facilitates the sampling and learning of any noisy version ${\mathbf{y}}_t$ directly. 

Specifically, for any given sample $(\mathbf{x},\mathbf{y})$ in $\mathcal{D}$, we introduce noise $\boldsymbol{\epsilon}_t$ sampled from a standard normal distribution $\mathcal{N}(0,\mathbf{I})$. And then randomly choose a time step $t$ from the uniform distribution and employ Eq. \ref{eq_ddpm_onestep_noising} to corrupt $\mathbf{y}$ as  
\begin{equation}\label{eq_ddpm_onestep_noising}
\mathbf{y}_t=\sqrt{\bar{\alpha}_t} \mathbf{y}_0+\sqrt{\left(1-\bar{\alpha}_t\right)} \boldsymbol{\epsilon}_t,
\end{equation}
where the added noise $\boldsymbol{\epsilon}_t$ is the prediction target of the model in denoising. 

For the denoising process, it reverses the corrupted sequence by estimating the true posterior distribution using a parametric Gaussian process as 
\begin{equation}\label{eq_ddpm_denoising}
p_\theta\left(\mathbf{y}_{t-1} \mid \mathbf{y}_t\right)=\mathcal{N}\left(\mathbf{y}_{t-1} ; \mu_\theta\left(\mathbf{y}_t, \mathbf{x}, t\right), \Sigma_\theta\left(\mathbf{y}_t, \mathbf{x}, t\right)\right),
\end{equation}

To handle conditional guidance from $\mathbf{x}$ during the denoising process, we use the Classifier-Free Guidance mechanism \cite{ho2022classifier} that jointly trains both conditional and unconditional models. Classifier-Free Guidance is able to handle some key challenges, such as the model's neglect of conditional factors and the issues stemming from imprecise score estimation and classifier gradients highlighted in \cite{beat2021gans}. This approach does not require training any additional models; it simply controls the proportion of conditional samples during DDPM training and, in the denoising process, combines the conditional and unconditional noise predictions through weighted summation. These modifications to the classic DDPM training and sampling process involve only a few lines of code, making it a low-complexity, high-quality solution for conditional generation. 

Specifically, in the training phase, a hyperparameter $p_{uncond}$ is introduced to determine to what extent the model should be trained without any condition (i.e., setting the condition input to $\varnothing$). This flexibility allows the model to learn robust representations in both conditioned and unconditioned contexts. In the sampling process, the intensity of conditional guidance is regulated by the weight $\omega$, which helps us achieve a refined score estimation. The resulting prediction noise is as 
\begin{equation}\label{eq_classifier_free_score}
    \tilde{\boldsymbol{\epsilon}}_{\theta}(\mathbf{y}_t,\mathbf{x},t) = (1 + \omega) \boldsymbol{\epsilon}_{\theta}(\mathbf{y}_t,\mathbf{x},t) - \omega \boldsymbol{\epsilon}_{\theta}(\mathbf{y}_t,t).
\end{equation}

The approach in Eq. \ref{eq_classifier_free_score} adeptly mitigates the issue of underestimating conditions while requiring only a single model for training. Classifier-Free Guidance strikes a balance between maintaining diversity and adhering to conditions. Although some diversity may be sacrificed, this trade-off is advantageous in the context of optimization, facilitating convergence toward high-quality solution distributions. Ultimately, we utilize the noise prediction from Eq. \ref{eq_classifier_free_score} to guide the generation process in optimization inference, resulting in target solutions. Using the predicted noise $\tilde{\boldsymbol{\epsilon}}_{\theta}$, the reconstruction function is defined as 
\begin{equation}\label{eq_reconstruct_func}
    \mathbf{y}_{t-1} = \frac{1}{\sqrt{\alpha_t}}(\mathbf{y}_t - \frac{1 - \alpha_t}{\sqrt{1 - \bar{\alpha}_t}} \tilde{\boldsymbol{\epsilon}}_{\theta}) + \frac{1 - \bar{\alpha}_{t-1}}{1 - \bar{\alpha}_t} \boldsymbol{\epsilon},\ \boldsymbol{\epsilon} \sim \mathcal{N}(0,\mathbf{I}).
\end{equation}

Under the guidance of Eq. \ref{eq_reconstruct_func}, the GDM can repeatedly sample from the target high-quality solution distribution $p(\mathbf{y}_0|\mathbf{x})$ based on the input $\mathbf{x}$, effectively approximating or reaching the optimal solution, as shown in Fig. \ref{fig_method_framework}. 

\subsection{Training and Sampling}

\begin{algorithm}[t]
\small
\caption{Training}
\label{alg:training} 
\KwIn{Dataset $\mathcal{D}$ with paired samples $(\mathbf{x}, \mathbf{y})$, Number of diffusion steps $T$, Noise schedule factors $(\alpha_t)_{t=1}^{T}$, Probability of unconditional sample $p_{uncond}$, Model parameters $\theta$;}
\KwOut{Trained model $\theta$;}

\Repeat{convergence}{
  Sample a data instance $(\mathbf{x}, \mathbf{y}) \in \mathcal{D}$\;
  Sample time step $t \sim \text{Uniform}(1, T)$\;
  Sample noise $\boldsymbol{\epsilon} \sim \mathcal{N}(0, \mathbf{I})$\;
  Compute noisy data $\mathbf{y}_t = \sqrt{\bar{\alpha}_t} \mathbf{y}_0 + \sqrt{1 - \bar{\alpha}_t} \boldsymbol{\epsilon}$\;
  Set the condition input $\mathbf{x}=\varnothing$ with probability $p_{uncond}$\;
  Compute model's predicted noise $\boldsymbol{\epsilon}_\theta(\mathbf{y}_t, \mathbf{x}, t)$\;
  Compute the loss:
  \[
  \mathcal{L}_\theta = \| \boldsymbol{\epsilon} - \boldsymbol{\epsilon}_\theta(\mathbf{y}_t, \mathbf{x}, t) \|^2
  \]
  Take gradient descent step on $\mathcal{L}_\theta$\;
}
\end{algorithm}

\begin{algorithm}[t]
\small
\caption{Sampling}
\label{alg:Sampling}
\KwIn{Number of diffusion steps $T$, Trained model $\theta$, Noise schedule factors $(\alpha_t)_{t=1}^{T}$, Input parameter $\mathbf{x}$;}
\KwOut{Generated solution $\mathbf{y}_0$;}

Sample initial noise $\mathbf{y}_T \sim \mathcal{N}(0, \mathbf{I})$\;
\For{$t = T, T-1, \dots, 1$}{
  Sample $\boldsymbol{\epsilon} \sim \mathcal{N}(0, \mathbf{I})$ if $t > 1$, else set $\boldsymbol{\epsilon} = 0$\;
  Update $\mathbf{y}_{t-1}$ as:
  \[
  \mathbf{y}_{t-1} = \frac{1}{\sqrt{\alpha_t}} \left( \mathbf{y}_t - \frac{1 - \alpha_t}{\sqrt{1 - \bar{\alpha}_t}} \tilde{\boldsymbol{\epsilon}}_\theta(\mathbf{y}_t, \mathbf{x}, t) \right) + \frac{1 - \bar{\alpha}_{t-1}}{1 - \bar{\alpha}_t} \boldsymbol{\epsilon}
  \]
}
\Return{$\mathbf{y}_0$}
\end{algorithm}

\subsubsection{Learning Objective and Loss Definition}
In the ideal scenario, we hope to use maximum likelihood estimation to make $D_{KL}(p(\mathbf{y}_0) || p_{\theta}(\mathbf{y}_0)) \approx 0$, that is, the difference between the target distribution $p(\mathbf{y}_0)$ and the distribution $p_{\theta}(\mathbf{y}_0)$ generated by the model is close to zero. However, since it is computationally challenging to directly solve $p(\mathbf{y}_0)$ and perform full trajectory integration, it must consider all possible reverse trajectories to obtain a more accurate estimate. In practical applications, alternative objective functions such as denoising score matching, $\boldsymbol{\epsilon}$-prediction, and the Evidence Lower Bound (ELBO) are commonly employed~\cite{kingma2024understanding}. We adopt the $\boldsymbol{\epsilon}$-prediction strategy to approximate $\nabla_{\mathbf{y}} \log p(\mathbf{y}_0 | \mathbf{y}_t, \mathbf{x}, t)$, thereby capturing gradient information across varying noise levels. The model learns to recover high-quality results from noisy data by minimizing the following loss function 
\begin{equation}\label{eq_condition_loss_typical}
    \mathcal{L}_{\theta} = \mathbb{E}_{\mathbf{y}, \mathbf{x}, \boldsymbol{\epsilon} \sim \mathcal{N}(0, \mathbf{I}), t}[\lVert \boldsymbol{\epsilon} - \boldsymbol{\epsilon}_{\theta}(\mathbf{y}_t, t, \mathbf{x}) \rVert^2].
\end{equation}

Eq. \ref{eq_condition_loss_typical} is iteratively used during training by repeatedly uniformly sampling $t$, detailed in \textbf{Algorithm \ref{alg:training}}. 


\subsubsection{Conditional Sampling}

The sampling process that leverages the classifier-free guidance generation approach is presented in \textbf{Algorithm \ref{alg:Sampling}}. This is a single-sampling process; in practice, parallel sampling can be performed to obtain multiple solutions within the same time consumption. Additionally, employing DDIM \cite{song2020ddim} with a few lines of code can reduce the total number of denoising steps, thereby increasing sampling efficiency. The training and sampling mechanism is adaptable to continuous, discrete, and structured optimization tasks. \vspace{-2mm}

\section{Results and Discussions}

\subsection{Experiments Settings}

\subsubsection{Problems}
We evaluate our GDM optimizer on three representative network optimization problems: computation offloading (CO)\cite{yang2021computation}, maximizing the sum rate across multiple channels (MSR)\cite{du2023beyond}, and maximizing the sum rate in a NOMA-UAV (NU) system~\cite{noma2021uav}. In the CO scenario, input parameters include network status and task information, with the output as the joint optimization of offloading decisions and resource allocation for an edge server. The objective is to minimize overall latency and power consumption, formulating a Mixed-Integer Non-Linear Programming (MINLP) problem. For the MSR scenario, inputs include channel gain parameters and a total power constraint, with the output providing the optimal power allocation across channels to maximize transmission rates, representing a standard convex optimization problem. Finally, in the NU problem, input data includes ground terminal coordinates and related network information, with the output as the joint optimization of UAV positioning and channel power allocation to maximize the sum rate, making it a hierarchical non-convex optimization problem. 

\subsubsection{Datasets}
We construct our dataset based on the problem formulations from previous studies \cite{yang2021computation, du2023beyond, noma2021uav}. The dataset consists of paired samples in the form $(\mathbf{x}, \mathbf{y})$, where $\mathbf{x}$ represents the input variables and $\mathbf{y}$ denotes the corresponding optimal solution. To obtain the optimal solutions, we apply an exhaustive search method. Specifically, for the CO problem, we conduct experiments involving joint optimization of offloading decisions and resource allocation in a setup with one edge server and three terminal nodes. For the MSR problem, we evaluate cases with $3$ and $80$ channels, with total available power of $10$W and $20$W, respectively. In the NU problem, we use a scenario involving three terminal nodes and one UAV node. The training dataset sizes for these three problems are $30,000$ samples for CO, $10,000$ samples for each of the MSR cases, and $10,000$ samples for NU. 
\subsubsection{Diffusion Models}
In the experiment, GDM optimizer adopts the classic DDPM \cite{ddpm2020ho} architechture and applies the conditional guidance mechanism \cite{ho2022classifier}. The neural network of our model is implemented using PyTorch. For the number of diffusion steps $T$ of the noising process in Eq. \ref{eq_ddpm_noising} we set an empirical value of $T=20$ after repeated experiments. The parameter $\omega$ in Eq. \ref{eq_classifier_free_score} is set to 500 by default in the experiments. The noise schedule $(\alpha_t)^T_{t=1}$ is constructed by the cosine schedule that has been widely adopted by researchers for diffusion models by default. And the unconditional training probability is set to $p_{uncond}=0.1$. We train the model uniformly for 200 epochs and use an Adam optimizer with an initial learning rate of 0.005. The learning rate is scaled down by $\gamma=0.1$ using the multi-step scheduler in the training process (milestones can be tuned). Exponential Moving Average (EMA) is used to improve model accuracy during training. The excessively large schedule factors for predicted noise in the early stage of the denoising process can lead to unconverged generation results so that normalization is introduced in the first $5$ steps of the denoising process. Given that repeated sampling intuitively yields better results, all experiments in this study adopt a single-sample solution generation approach. This allows us to focus on feasibility validation and exploring the impact of key variables. 
\subsubsection{Baselines}
We employ three representative baseline methods: (1) Gradient Descent (GD) \cite{optimization2020liu} as a representative numerical method; (2) Multi-Task Feedforward Neural Network (MTFNN) \cite{yang2021computation} as a representative of supervised deep learning; and (3) Proximal Policy Optimization (PPO) \cite{schulman2017ppo} as a representative of reinforcement learning. 
\subsubsection{Metrics}
We set the ratio of the objective function value corresponding to the output solution $\mathbf{y}_0$ and the ground truth $\mathbf{y}_{true}$ as the evaluation criterion, i.e. $\frac{f(\mathbf{x},\mathbf{y}_0)}{f(\mathbf{x},\mathbf{y}_{true})}$, named Exceed\_ratio. The closer the ratio is to 1, the better.

\subsection{Experiments on Diffusion Model Settings}
To provide a correct and efficient configuration for using GDM as a network optimizer, we first analyze the impact of the hyperparameters and external settings of the proposed GDM optimizer to show the potential features. 

\subsubsection{Condition with Objective and Constraint Terms}
\begin{figure}[t]
\setlength{\abovecaptionskip}{-1.2mm}
\centerline{\includegraphics[width=3.4in]{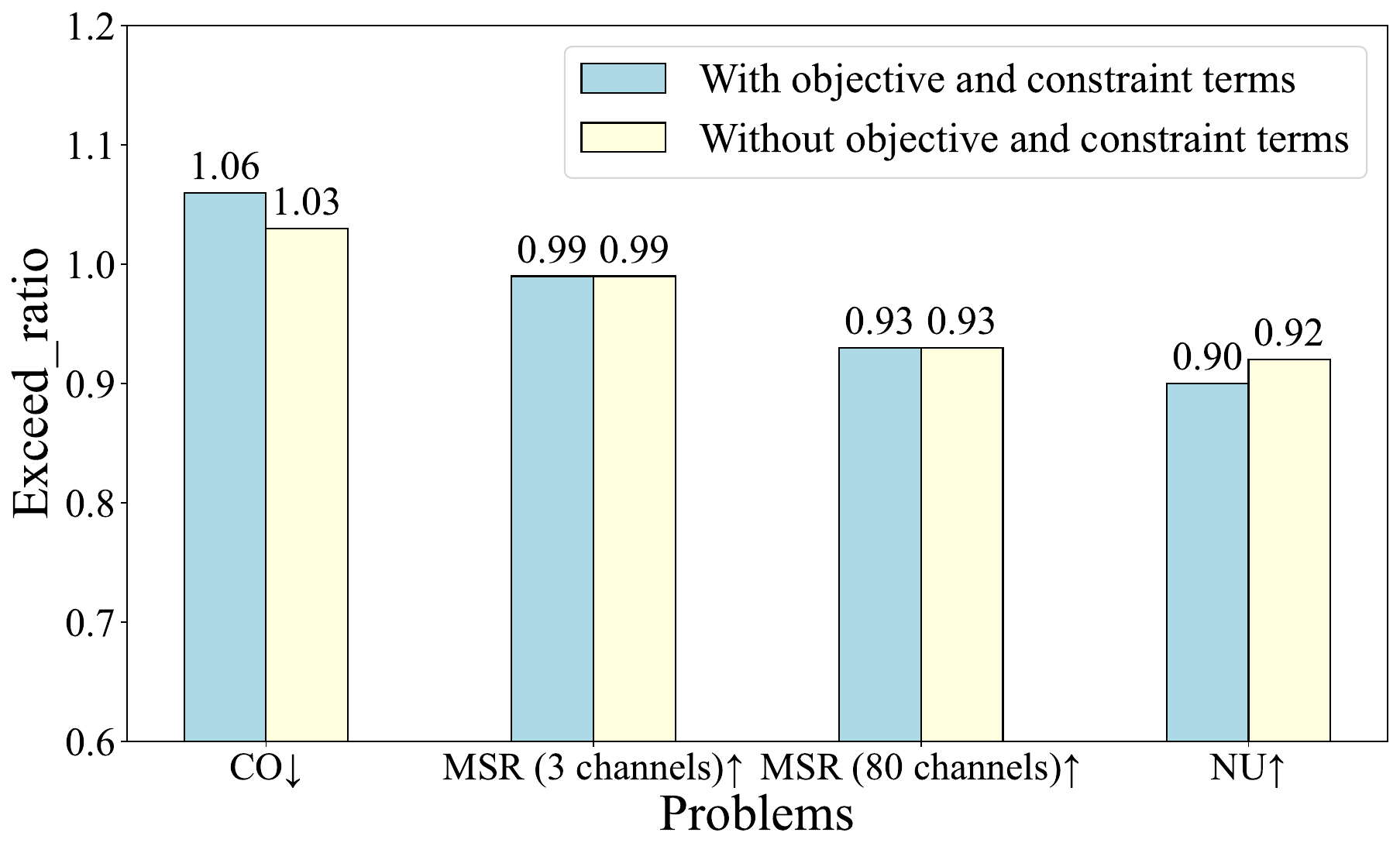}}
\caption{\small The effects of objective and constraint terms within the conditions.}
\label{fig_objective_term_comp}
\vspace{-4mm}
\end{figure}

Instructing the quality of the current solution to the model can provide more meaningful information, intuitively it can improve the learning effect. In the experiments, we incorporate the objective function value and the Lagrangian constraint function value of the solution into the condition as model input. But in fact, ablation experiments in Fig. \ref{fig_objective_term_comp} show that the model performance has not been significantly improved or even slightly worse after incorporation. We believe that it may be because the objective and constraint values are more difficult to understand after being numerically encoded, which has almost no distribution properties. 

\subsubsection{Strength of Condition}

\begin{figure}[t]
\setlength{\abovecaptionskip}{-1.2mm}
\centerline{\includegraphics[width=3.4in]{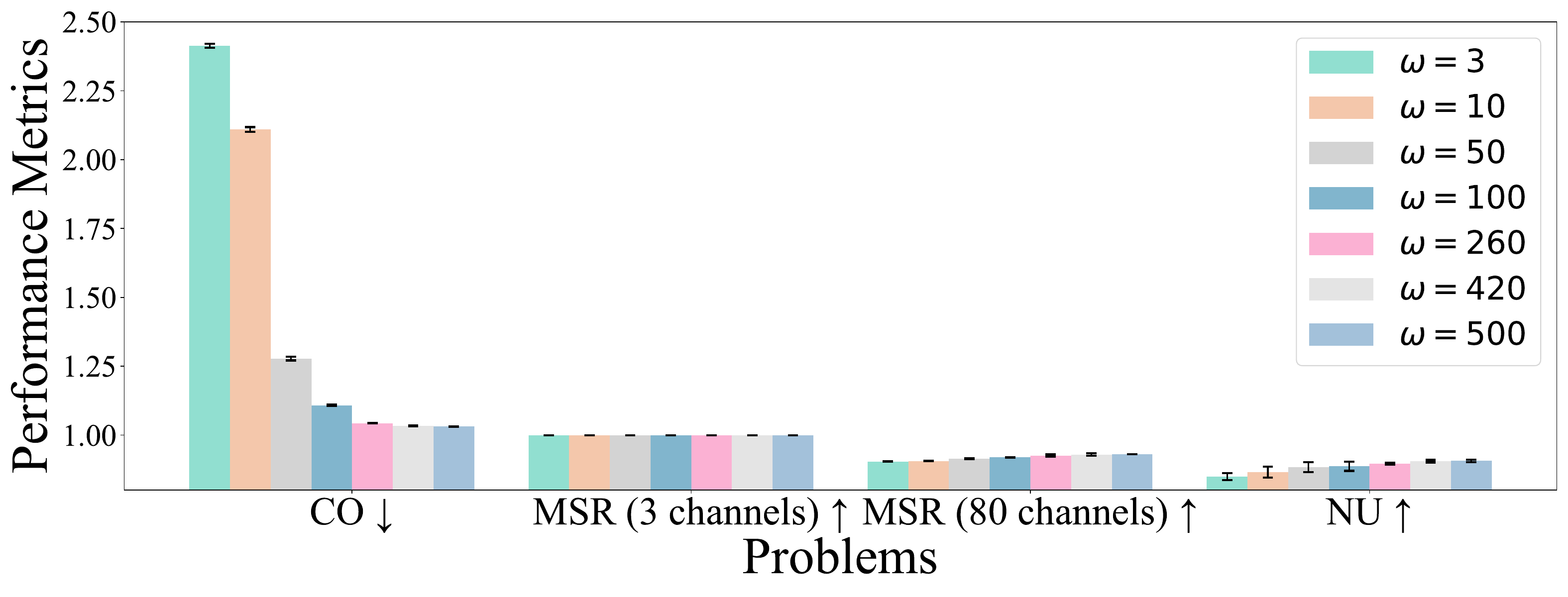}}
\caption{\small The effects of different condition strengths. Each bar includes upper and lower limit lines to indicate the range of variation observed across repeated experiments.}
\label{fig_omega_settings}
\vspace{-4mm}
\end{figure}

\begin{figure}[t]
\setlength{\abovecaptionskip}{-1.2mm}
\centerline{\includegraphics[width=3.35in]{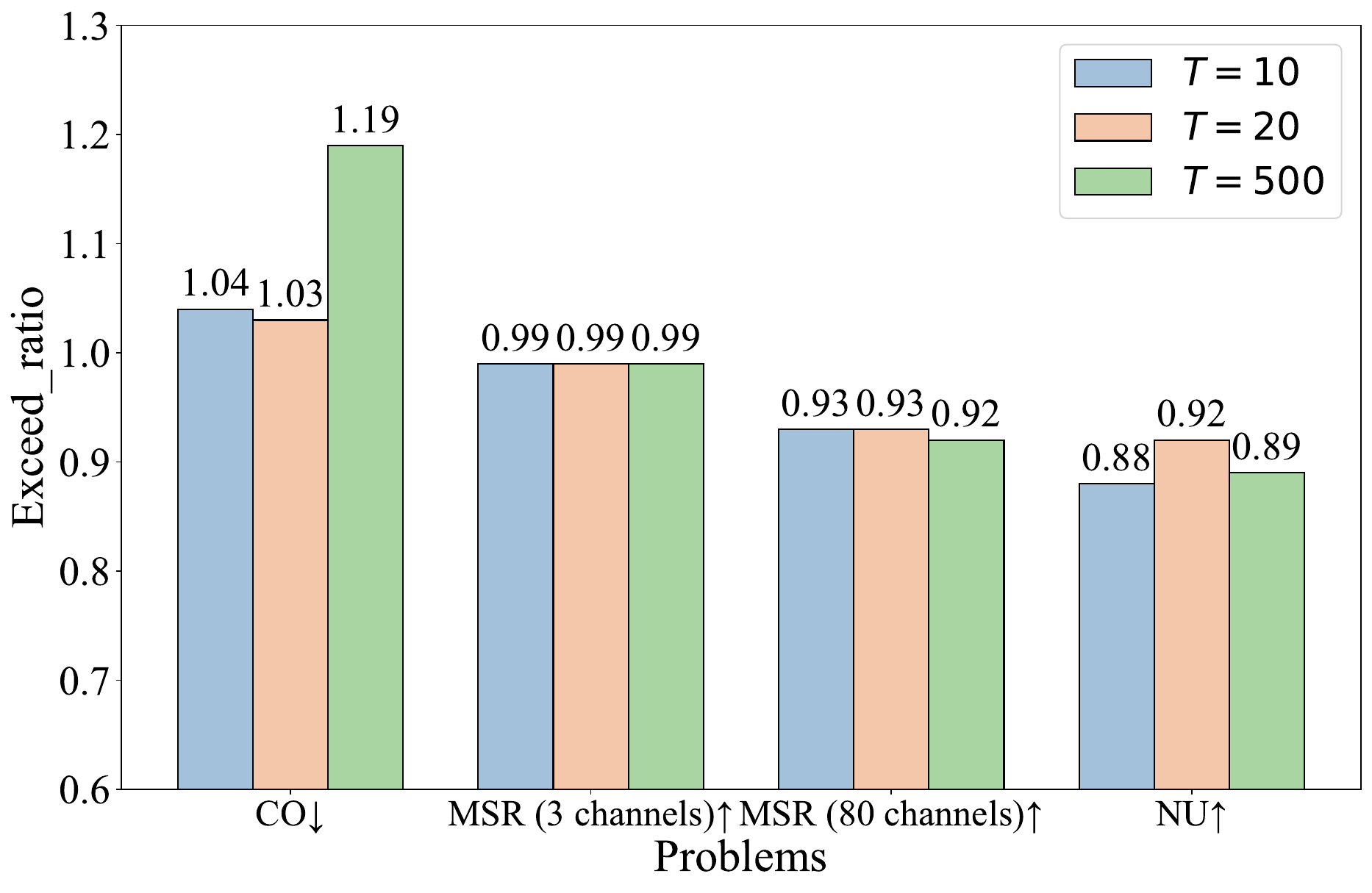}}
\caption{\small The effects of the number of diffusion steps $T$.}
\label{fig_T_settings}
\vspace{-3mm}
\end{figure}

\begin{figure}[t]
\setlength{\abovecaptionskip}{-1.2mm}
\centerline{\includegraphics[width=3.4in]{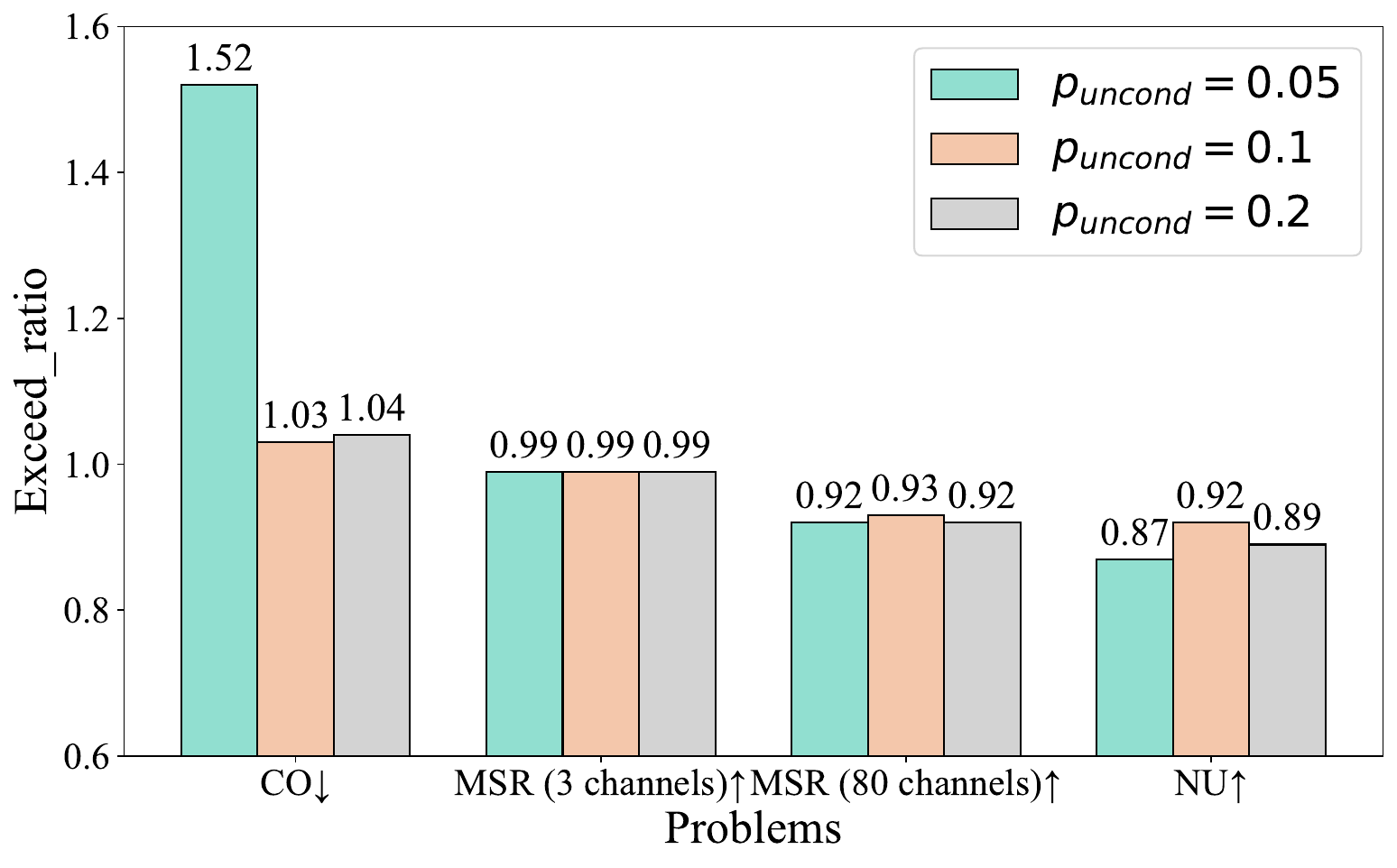}}
\caption{\small The effects on optimization performance of $p_{uncond}$.}
\label{fig_p_uncond_obj}
\vspace{-3mm}
\end{figure}

\begin{figure}[t]
\setlength{\abovecaptionskip}{-1.2mm}
\centerline{\includegraphics[width=3.4in]{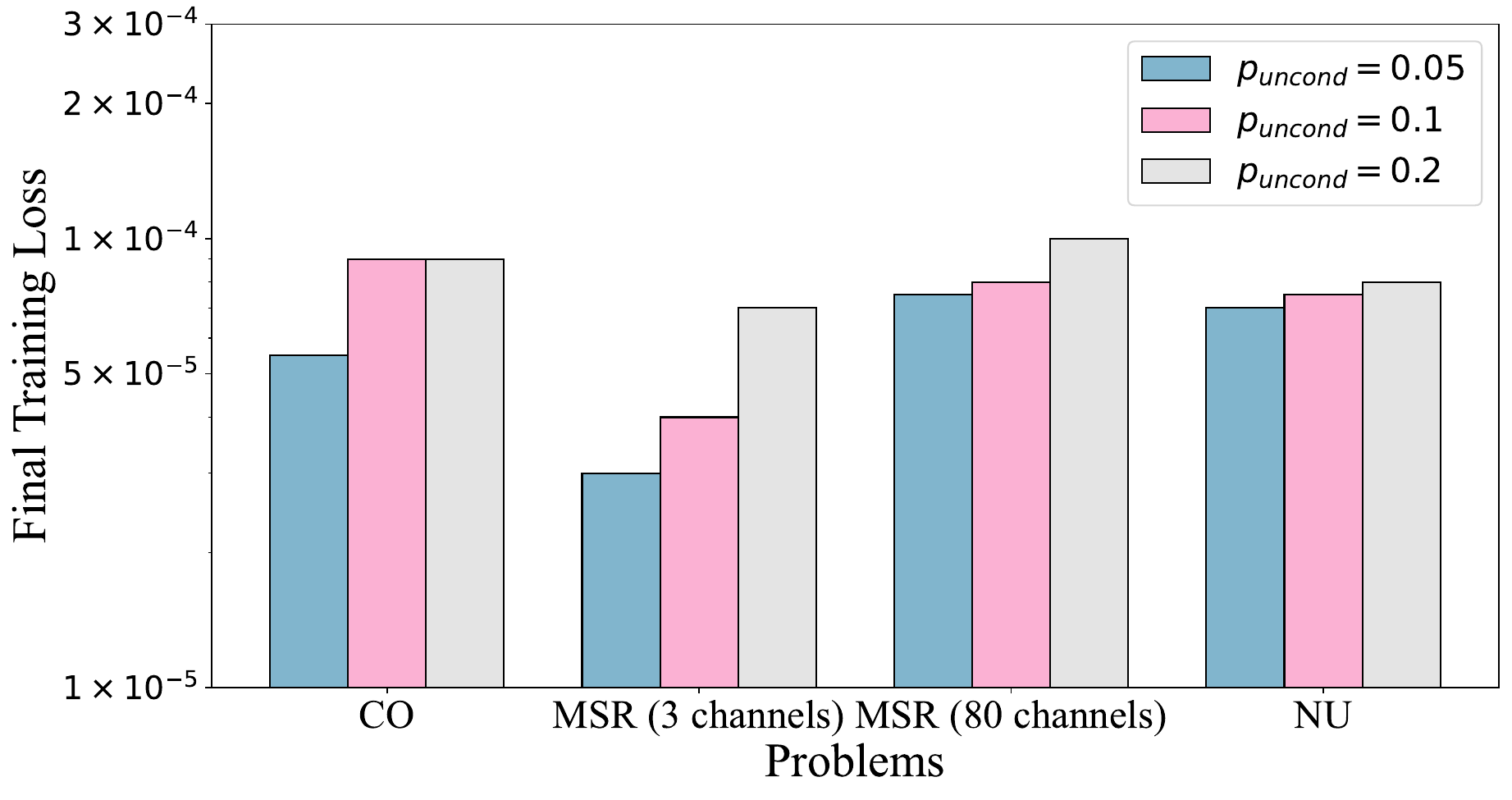}}
\caption{\small The effects on final training loss of $p_{uncond}$.}
\label{fig_p_uncond_loss}
\vspace{-3mm}
\end{figure}

For the influence of the conditional strength parameter $\omega$, conventional $w$ is set to less than 4 in image generation \cite{ho2022classifier}. However, as shown in Fig. \ref{fig_omega_settings}, $\omega$ has a great impact in the two cases of CO and NU, but has almost no impact in the two cases of MSR. We found that the magnitude of the impact of $\omega$ may be related to the complexity of the dataset and the problem itself. The more complex the problem, the greater the impact of $\omega$. More in-depth research on these influencing factors needs to be carried out.

\subsubsection{Number of Diffusion Steps}

For the number of diffusion steps $T$. As shown in Fig. \ref{fig_T_settings}, $T$ that is too large or too small will both make poor learning results. We believe that it is because $T$ is too small to ensure the Gaussian result distribution of the forward process causing the wrong reverse process, while $T$ that is too large will bring more knowledge to be learned, resulting in insufficient training epochs and model capacity. A large number of diffusion steps may also result in redundant denoising costs. 

\subsubsection{Unconditional Probability of Sample in Training}

Experiments show that $p_{uncond}$ may determine the lower bound of the training loss, and the smaller the $p_{uncond}$, the smaller the magnitude of the final training loss, as shown in Fig. \ref{fig_p_uncond_loss}. However, $p_{uncond}=0.1$ achieves the best model performance in practice, as shown in Fig. \ref{fig_p_uncond_obj}. The lower bound of training loss decreases as the unconditional probability decreases because the learning objective of the model approaches pure conditional inference as the unconditional probability decreases. For the performance decline, we believe it is because the model cannot fully learn the conditional guidance information when $p_{uncond}$ is too large. If $p_{uncond}$ is too small, the difference between unconditional and conditional predictions will be indiscernible, and effective weighted condition guidance cannot be achieved.

\subsection{Denoising Process}

To provide a more intuitive analysis of the solution generation process of GDM as an optimizer, we select several input samples from each of the three problems and visualize the changes in the objective function values of their solutions as denoising progresses over the denoising steps. 

\begin{figure}[t]
\setlength{\abovecaptionskip}{-1.2mm}
\centerline{\includegraphics[width=3.25in]{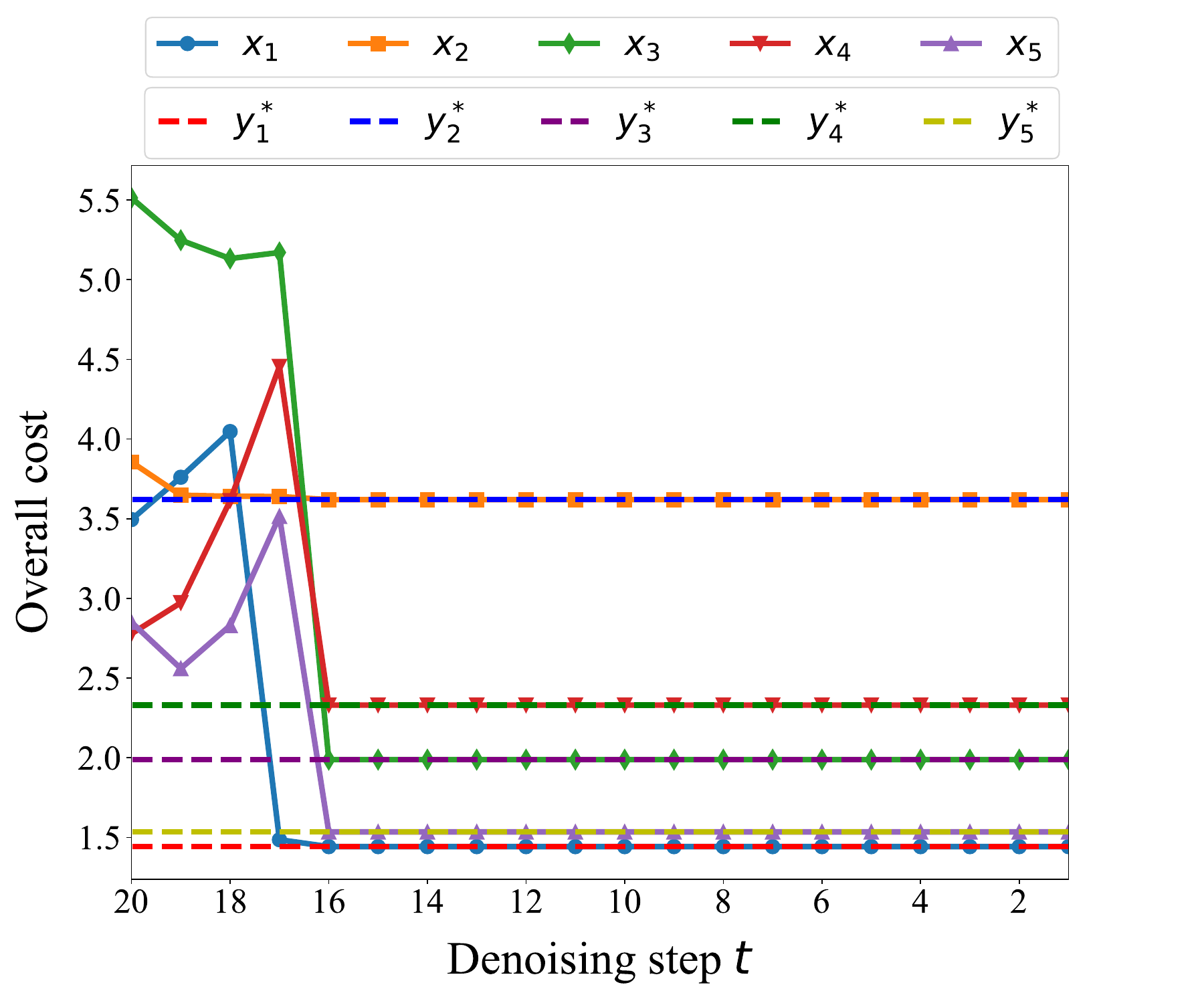}}
\caption{\small Objective values of the denoising process for 5 different samples of the CO problem and their corresponding optimal solution benchmarks. For example, $\mathbf{x}_1$ represents a feature input, with its solution generation process depicted using a solid line. $\mathbf{y}_1^*$ denotes the ground truth solution, represented by a dashed line indicating its corresponding optimal cost.}
\label{fig_co_denoising}
\vspace{-4mm}
\end{figure}

\begin{figure}[t]
\setlength{\abovecaptionskip}{-1.2mm}
\centerline{\includegraphics[width=3.2in]{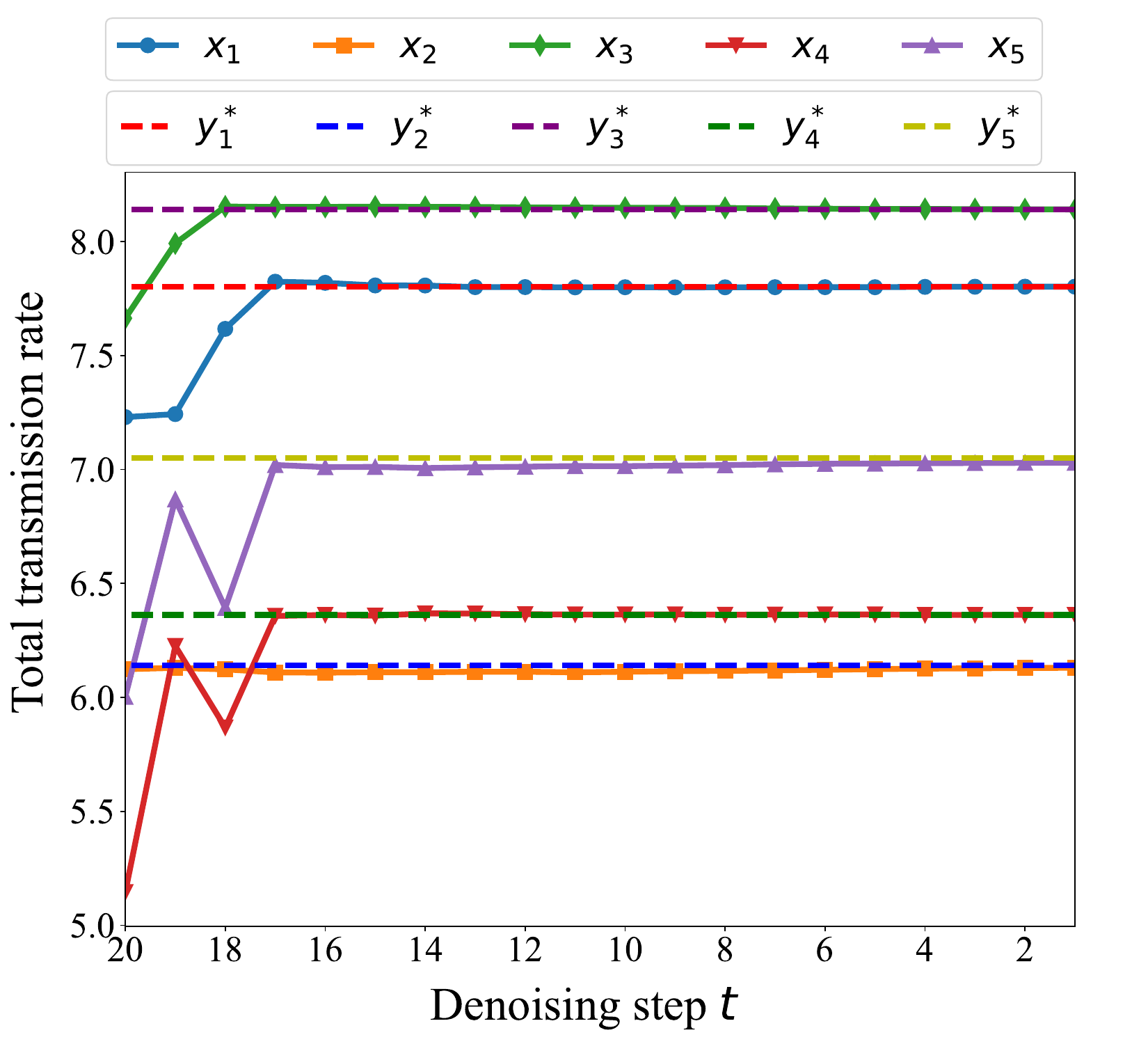}}
\caption{\small Objective values of the denoising process for 5 different samples of the MSR problem and their corresponding optimal solution benchmarks. The meaning of $\mathbf{x}$ and $\mathbf{y}$ are consistent with those in Fig. \ref{fig_co_denoising}.}
\label{fig_msr_denoising}
\vspace{-4mm}
\end{figure}

\begin{figure}[h]
\setlength{\abovecaptionskip}{-1.2mm}
\centerline{\includegraphics[width=3.25in]{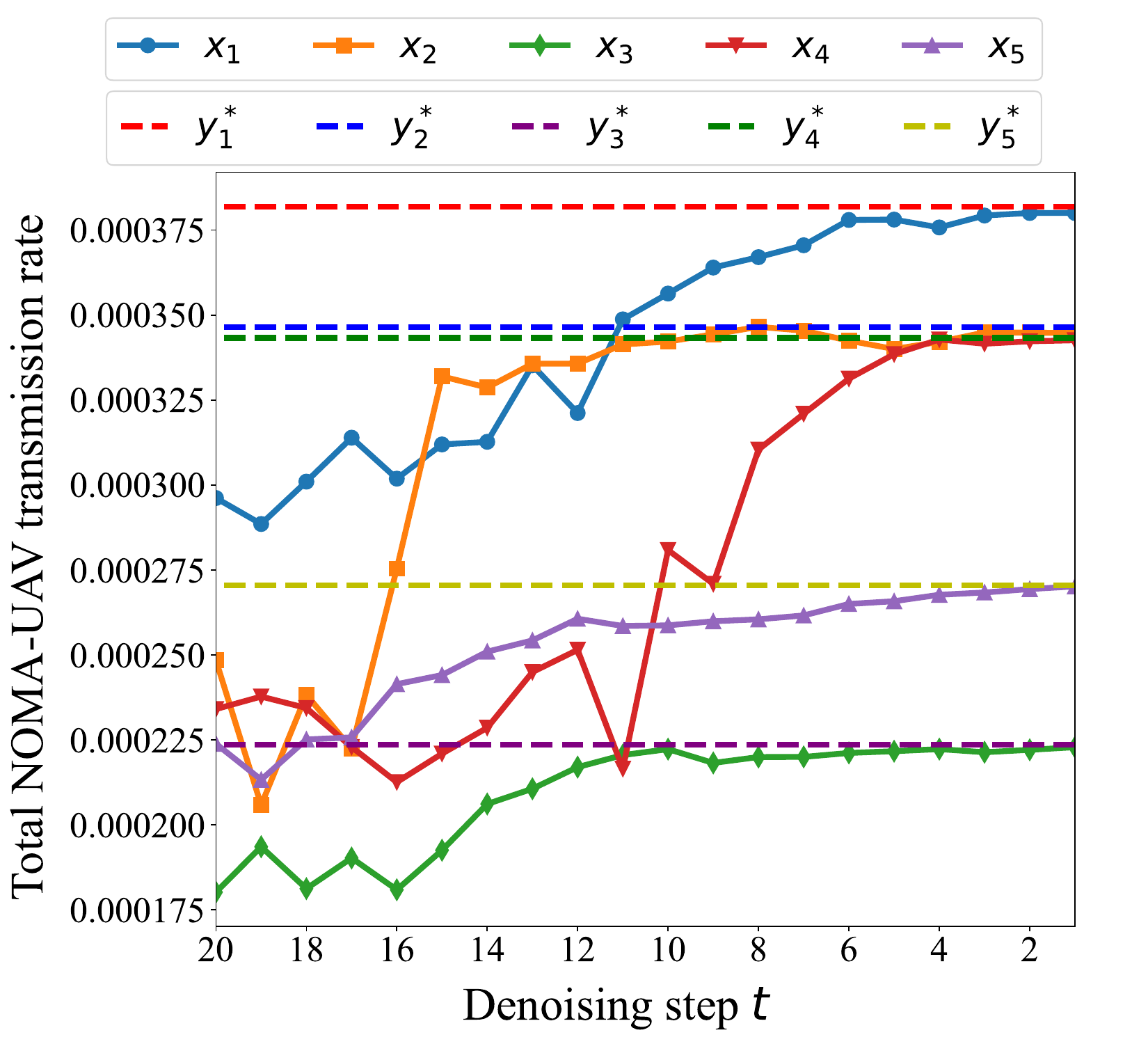}}
\caption{\small Objective values of the denoising process for 5 different samples of the NU problem and their corresponding optimal solution benchmarks. The meaning of $\mathbf{x}$ and $\mathbf{y}$ are consistent with those in Fig. \ref{fig_co_denoising}.}
\label{fig_nu_denoising}
\vspace{-4mm}
\end{figure}

\begin{figure}[t]
\setlength{\abovecaptionskip}{-1.2mm}
\centerline{\includegraphics[width=3.25in]{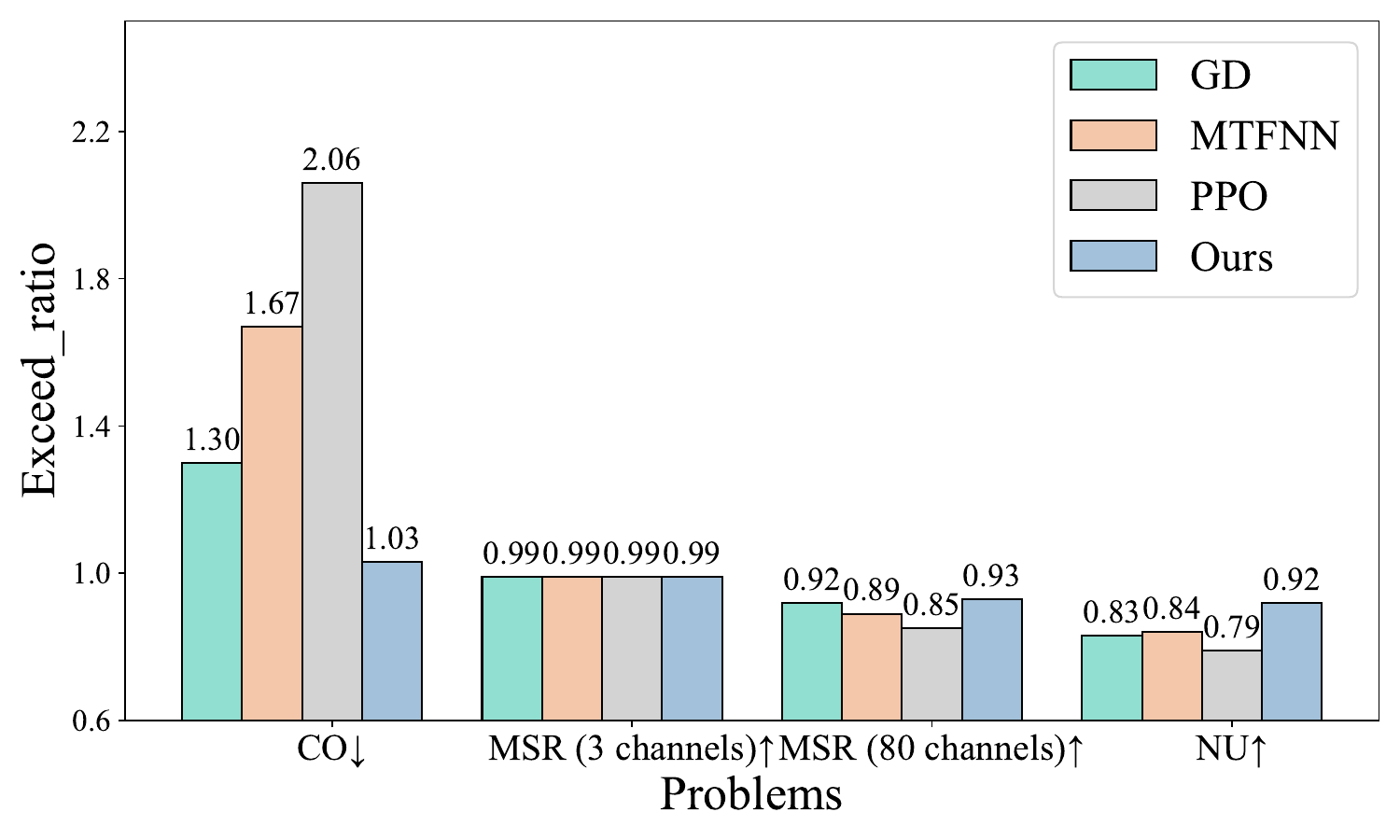}}
\caption{\small Optimization comparisons between our GDM optimizer and baselines.}
\label{fig_opt_performance_comp}
\vspace{-4mm}
\end{figure}

\subsubsection{CO Problem}
As shown in Fig. \ref{fig_co_denoising}, the horizontal axis represents the denoising time steps, while the vertical axis shows the global cost of computation offloading, which we aim to minimize. Each colored line corresponds to the denoising process of a sample's solution, with each point representing the global cost at a specific time step. The dashed lines indicate the optimal total cost for each input's optimal solution; for example, the red dashed line marks the optimal cost of the blue line's input $\mathbf{x}_1$ and its optimal solution $\mathbf{y}_1^*$. 

The figure shows a fluctuating trend in solution generation, highlighting the nature of denoising toward a high-quality solution distribution rather than pure gradient descent. The final denoising outcomes closely approach or reach the optimal solution, demonstrating stable convergence. Moreover, convergence is achieved in under five steps, primarily due to the problem's relatively low scale and complexity, as well as a high conditioning strength factor $\omega$. 

\subsubsection{MSR Problem}

Similarly, in Fig. \ref{fig_msr_denoising}, the vertical axis shows the total transmission rate, which we aim to maximize. Other settings of Fig. \ref{fig_msr_denoising} follow the same format as Fig. \ref{fig_co_denoising}. Since the high-quality solution space of individual inputs to this problem occupies a large area of the feasible solution space, the initial noise solution has performance close to the optimal solution, such as $\mathbf{x}_2$ in Fig. \ref{fig_msr_denoising}. In addition, we observe that MSR converges faster than CO, which aligns with the fact that MSR, as a convex optimization problem, has a lower complexity than the MINLP required for CO. 

\subsubsection{NU Problem}

In Fig. \ref{fig_nu_denoising}, the vertical axis represents the total transmission rate in the NOMA-UAV scenario, which is the target for maximization. The settings in Fig. \ref{fig_nu_denoising} follow the same format as in Fig. \ref{fig_msr_denoising}. For the NU problem, we observe a slower convergence compared to both CO and MSR, requiring at least ten denoising steps. This reflects the high complexity of this hierarchical, non-convex optimization problem. 

\subsubsection{Summary}
In summary, our GDM optimizer consistently demonstrated stable sampling of high-quality solution distributions across all three problems, effectively converging toward the optimal solution. Additionally, we observed that as the complexity or scale of the problem increases, the number of denoising steps required for convergence also rises. The complexity of the problems MSR, CO, and NU increases progressively from easy to difficult, leading to slower solution generation convergence under the same model settings. This is related to the size and complexity of the solution space, as non-convex and truncated mixed-integer problems present significantly greater challenges for learning and memorization. 

\subsection{Optimization Performance}


As shown in Fig. \ref{fig_opt_performance_comp}, our proposed GDM optimizer consistently outperforms the baseline methods, with the exception of the 3-channel MSR scenario, where results are comparable. The poor performance of GD in the CO and NU problems is primarily due to the mixed-integer and non-convex characteristics of these problems; even when relaxed using the Lagrange multiplier method, gradient descent tends to get trapped in local optima. Further, transforming such complex problem properties into convex forms would require specialized algorithmic designs, underscoring the advantage of the GDM optimizer’s robustness to different objective function characteristics. The deficiencies of MTFNN and PPO, as neural network-based methods, stem not only from the prediction error discussed in Sec. \ref{sec_gen_advan} but also from fundamental limitations: the mixed-integer nature of the CO problem makes it difficult to capture the input-to-optimal-solution relationship using differentiable models, while the non-convex nature of the NU problem presents similar challenges. This highlights the strength of generative models in learning high-quality solution distributions to overcome such complexities. In summary, our GDM optimizer demonstrates exceptional performance across problems of increasing complexity and scale, validating its theoretical and empirical efficacy. \vspace{-2mm}

\section{Conclusion}
In this work, we bridge the theoretical gaps in GDM as an emerging network optimizer and provide further empirical guidance. This novel network optimization approach holds the potential to deliver significant improvements in both efficiency and quality for IoT networks. Our study has provided a concise theoretical demonstration of the inherent advantages of generative models over discriminative models in optimization solving. Specifically, GDMs offer a promising approach by learning the underlying high-quality solution distribution of feasible space and enabling repeated sampling from this distribution, which improves their ability to handle complex solution spaces and achieve optimal results. By capturing global solution structures rather than directly mapping inputs to outputs, GDMs address the inherent limitations of discriminative models. Our work has leveraged these strengths of GDMs, demonstrating their capacity to not only enhance the flexibility and accuracy of network optimization but also streamline the optimization process across alternative scenarios. 

For future work, we aim to address the current GDM optimizer’s reliance on supervised training by further advancing both theory and practice to design methods that release the high dependency on data. 


\ifCLASSOPTIONcaptionsoff
  \newpage
\fi



%

%

\bibliographystyle{IEEEtran}  
\bibliography{iotj}




\end{document}